\documentclass[lettersize,journal]{IEEEtran}
\usepackage{algorithmic}
\usepackage{array}
\usepackage[caption=false,font=normalsize,labelfont=sf,textfont=sf]{subfig}
\usepackage{textcomp}
\usepackage{stfloats}
\usepackage{url}
\usepackage{verbatim}
\def\BibTeX{{\rm B\kern-.05em{\sc i\kern-.025em b}\kern-.08em
    T\kern-.1667em\lower.7ex\hbox{E}\kern-.125emX}}
\usepackage{balance}

\usepackage{graphics} 
\usepackage{epsfig} 
\usepackage{times} 
\usepackage{amsmath} 
\usepackage{amssymb}  
\usepackage{amsfonts,amssymb}
\usepackage{bm}
\usepackage{graphicx}
\usepackage{float} 
\usepackage[outdir=./]{epstopdf}
\usepackage{color}
\usepackage[dvipsnames]{xcolor}
\usepackage{setspace}
\usepackage{multirow}
\usepackage{booktabs}

\makeatletter
\renewcommand{\maketag@@@}[1]{\hbox{\m@th\normalsize\normalfont#1}}%
\makeatother

\title{
Interactive Motion Planning for Human-Robot Collaboration Based on Human-Centric Configuration Space Ergonomic Field
}
\author{}

\author{Chenzui Li, Yiming Chen, Xi Wu, Tao Teng, Sylvain Calinon, Darwin Caldwell, \IEEEmembership{Fellow, IEEE}, \\ and Fei Chen, \IEEEmembership{Senior Member, IEEE}
\thanks{This work was supported in part by the Research Grants Council of the Hong Kong SAR under Grant 24209021, 14222722, 14211723 and C7100-22GF and in part by InnoHK of the Government of Hong Kong via the Hong Kong Centre for Logistics Robotics. (\textit{Corresponding author: Fei Chen)}}
\thanks{Chenzui Li, Yiming Chen, Xi Wu, Tao Teng, and Fei Chen are with the Department of Mechanical and Automation Engineering, T-Stone Robotics Institute, The Chinese University of Hong Kong, Hong Kong (e-mail: {czli@mae.cuhk.edu.hk, ymchen@mae.cuhk.edu.hk, xwu@mae.cuhk.edu.hk, tao.teng@ieee.org, f.chen@ieee.org}).}%
\thanks{Sylvain Calinon is with the Idiap Research Institute, Martigny, Switzerland (e-mail: {sylvain.calinon@idiap.ch}).}
\thanks{Darwin Caldwell is with the Department of Advanced Robotics, Istituto Italiano di Tecnologia, 16163 Genoa, Italy (e-mail: {darwin.caldwell@iit.it}).}

}

\begin{document}

\maketitle

\begin{abstract}
Industrial human-robot collaboration requires motion planning that is collision-free, responsive, and ergonomically safe to reduce fatigue and musculoskeletal risk. We propose the Configuration Space Ergonomic Field (CSEF), a continuous and differentiable field over the human joint space that quantifies ergonomic quality and provides gradients for real-time ergonomics-aware planning. An efficient algorithm constructs CSEF from established metrics with joint-wise weighting and task conditioning, and we integrate it into a gradient-based planner compatible with impedance-controlled robots. In a 2-DoF benchmark, CSEF-based planning achieves higher success rates, lower ergonomic cost, and faster computation than a task-space ergonomic planner. Hardware experiments with a dual-arm robot in unimanual guidance, collaborative drilling, and bimanual co-carrying show faster ergonomic cost reduction, closer tracking to optimized joint targets, and lower muscle activation than a point-to-point baseline. CSEF-based planning method reduces average ergonomic scores by up to 10.31\% for collaborative drilling tasks and 5.60\% for bimanual co-carrying tasks while decreasing activation in key muscle groups, indicating practical benefits for real-world deployment.
\end{abstract}


\section{Introduction} \label{introduction}

Ergonomics plays a critical role in human-robot collaboration (HRC), ensuring human comfort and safety during collaborative tasks~\cite{lorenzini2023ergonomic}. As robots increasingly share workspaces in manufacturing, healthcare, and domestic environments, consideration of human posture and movement becomes essential for collaboration. Poor ergonomics can cause musculoskeletal disorders (MSDs) and reduced productivity~\cite{da2010risk}, whereas ergonomic optimization improves performance, satisfaction, and long-term health. Despite advances in robotics, many systems still overlook physical ergonomics, leaving a gap between technical capability and human-centric deployment.

Physical human-robot interaction (pHRI) introduces challenges for embedding ergonomics into motion planning, as robots should adapt continuously to human movements while maintaining ergonomic constraints. Postural optimization is commonly used and has been applied to tasks such as handover~\cite{busch2017postural,ovur2025optimising}, co-manipulation~\cite{van2020predicting,marin2018optimizing}, and co-carrying~\cite{kim2017anticipatory}. However, current ergonomic assessments are discrete and require modification for continuous planning~\cite{yazdani2022dula}. Moreover, many planners prioritize robot efficiency and obstacle avoidance over ergonomic benefit, leading to technically valid but ergonomically suboptimal interactions.

\begin{figure}[!t]
\centering
\includegraphics[width=1\linewidth]{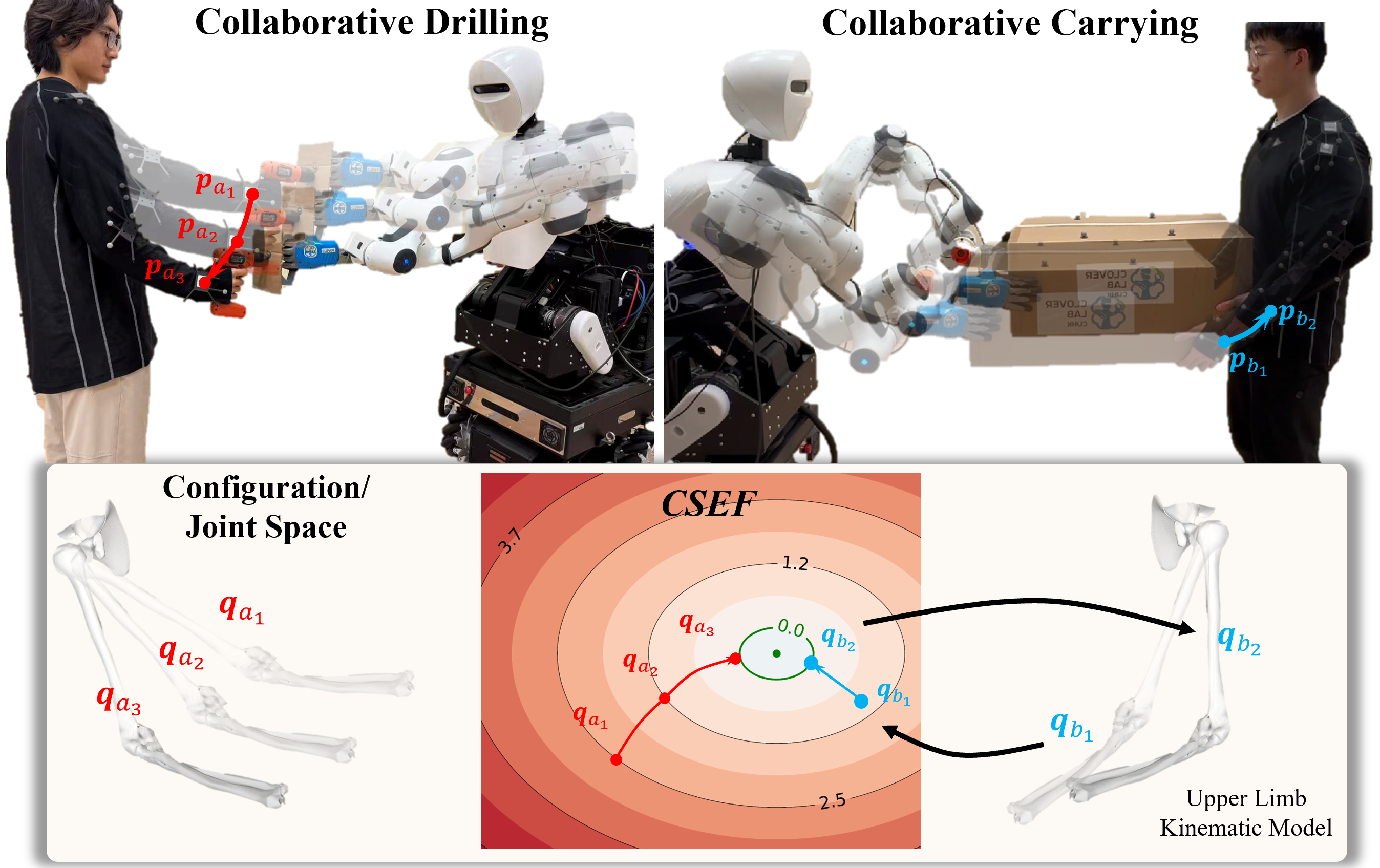} 
\caption{Illustration of the potential application with the CSEF for motion planning in the human-robot collaborative drilling and carrying tasks. (We declare that all the individuals in this figure and any other figure in this manuscript gave permission for the use of their images.)}
\label{Fig.cover}
\vspace{-4mm}
\end{figure}

Signed Distance Fields (SDFs) are powerful implicit representations in robotics, offering geometric information that integrates well with control, optimization, and learning~\cite{oleynikova2016signed}. Their differentiability and unit norm gradient properties suit gradient-based planning, collision avoidance, and shape modeling~\cite{liu2023collision}. Recent work extends SDF to Configuration Space Distance Fields that measure angular distances in joint space while preserving Euclidean properties~\cite{li2024configuration}. Building on this, we introduce Configuration Space Ergonomic Fields (CSEFs), which transpose signed fields from geometric distance to ergonomic assessment. Specifically, CSEF evaluates human joint configurations, yielding a continuous scalar field where lower values indicate more comfortable postures (Fig.~\ref{Fig.cover}).

In this paper, we define CSEF as a continuous and differentiable scalar field over human configurations, analyze its smoothness and gradient informativeness under joint limits and kinematic feasibility, and present an efficient algorithm to construct and update it from established ergonomic metrics with joint-wise weighting and subject-specific models. Building on CSEF, we develop a gradient-based interactive planner that guides humans toward ergonomically favorable postures. We validate the approach across unimanual guidance, collaborative drilling, and bimanual co-carrying, comparing against a task-space ergonomics-based planner in simulation and a point-to-point (PTP) baseline on hardware.

The primary contributions of this work are:
\begin{itemize}
\item [1)]
A configuration space ergonomic field for human-centric physical ergonomics representation that provides smooth gradients for motion planning.
\item [2)]
An interactive framework that unifies ergonomic metrics with a human skeletal model to generate CSEF, and embeds it into interactive motion planning and impedance-based robot trajectory generation for real-time human–robot collaboration.
\item [3)]
Industrially oriented validation on a bimanual collaborative robot by representative shop-floor tasks such as unimanual guidance, collaborative drilling, and bimanual co-carrying, demonstrating improved ergonomic outcomes and obvious muscle activation reduction.
\end{itemize}

\section{Related Works} \label{related works}
\subsection{Physical Ergonomics Assessment for Postural Optimization}
Effective postural optimization in HRC requires credible models of human comfort and risk. Practice-oriented tools such as RULA~\cite{mcatamney1993rula}, REBA~\cite{hignett2000rapid}, the Strain Index~\cite{steven1995strain}, and OWAS~\cite{gomez2017musculoskeletal} offer expert-driven, discrete risk scores with strong empirical validation and high interpretability. Complementary computational proxies tailored to interactive settings model muscle fatigue~\cite{peternel2019selective}, joint overload~\cite{lorenzini2019new}, and human comfort~\cite{figueredo2020human}.

For postural optimization in HRC, RULA and REBA provide posture-centric quantitative scores but are discrete or piecewise and not natively differentiable, hindering gradient-based optimization. To address this, DULA/DEBA were introduced with continuous, differentiable surrogates of RULA/REBA that retain semantics and enable real-time gradient-based assessment with near-equivalent accuracy~\cite{yazdani2022dula}. NeuroErgo uses a neural approximation of tabular assessors (instantiated for REBA), improves score prediction, and is directly differentiable for gradient-based optimization~\cite{nejadasl2022neuroergo}. A comfort predictor combining a muscle-activation metric with a REBA-derived ergonomic metric enables fast queries via an offline voxelized workspace map for unimanual collaboration~\cite{figueredo2020human}. Similarly, upper-limb posture can be optimized by converting RULA to a continuous cost and combining it with human force manipulability, then steering humans via robot end-effector pose adjustments for bimanual collaboration~\cite{li2025integrating}.


\subsection{Ergonomic-Based Motion Planning}
While prior HRC studies improved physical ergonomics, they often underemphasize motion and path planning, namely how the robot should move to meet human ergonomic objectives. A unified framework infers human dynamics and intentions such as joint overload, target selection, handedness, and worker movement, and adjusts robot behavior to minimize joint loading~\cite{kim2021human}. Another planner predicts and reduces muscular effort from kinematics and task forces~\cite{figueredo2021planning}. Both reduce torque and effort but lack real-time, continuous motion planning during collaboration. More recently, an ergonomically optimized path planner guides workers along ergonomic paths using Cartesian planning and a REBA-based cell decomposition~\cite{merikh2024ergonomically}. However, task-space cell decomposition overlooks joint-space redundancy, and demonstrations are limited to simple two-dimensional settings with a largely static human and fixed lower limbs. 

In exoskeleton scenarios, a trajectory planning framework was introduced for wearable supernumerary arms to minimize user muscle load during operation of a wearable robotic forearm \cite{vatsal2021biomechanical}. The method incorporates a high-fidelity human arm model to derive a biomechanical cost term, which is integrated into the motion planner’s objective. An ergonomic-driven multi-objective trajectory optimization framework for human workers was proposed in \cite{gomes2021multi}. It formulates path and posture planning using human kinematic and kinetic states as objectives and constraints, jointly considering multiple ergonomics risk metrics across different body regions.

\section{Configuration Space Ergonomic Field (CSEF)} \label{CSEF}
In this section, we introduce CSEF and explore its properties. We then present an efficient algorithm to compute CSEF, as well as the potential application of CSEF to handle HRC.

\begin{figure}[!t]
\centering
\includegraphics[width=1\linewidth]{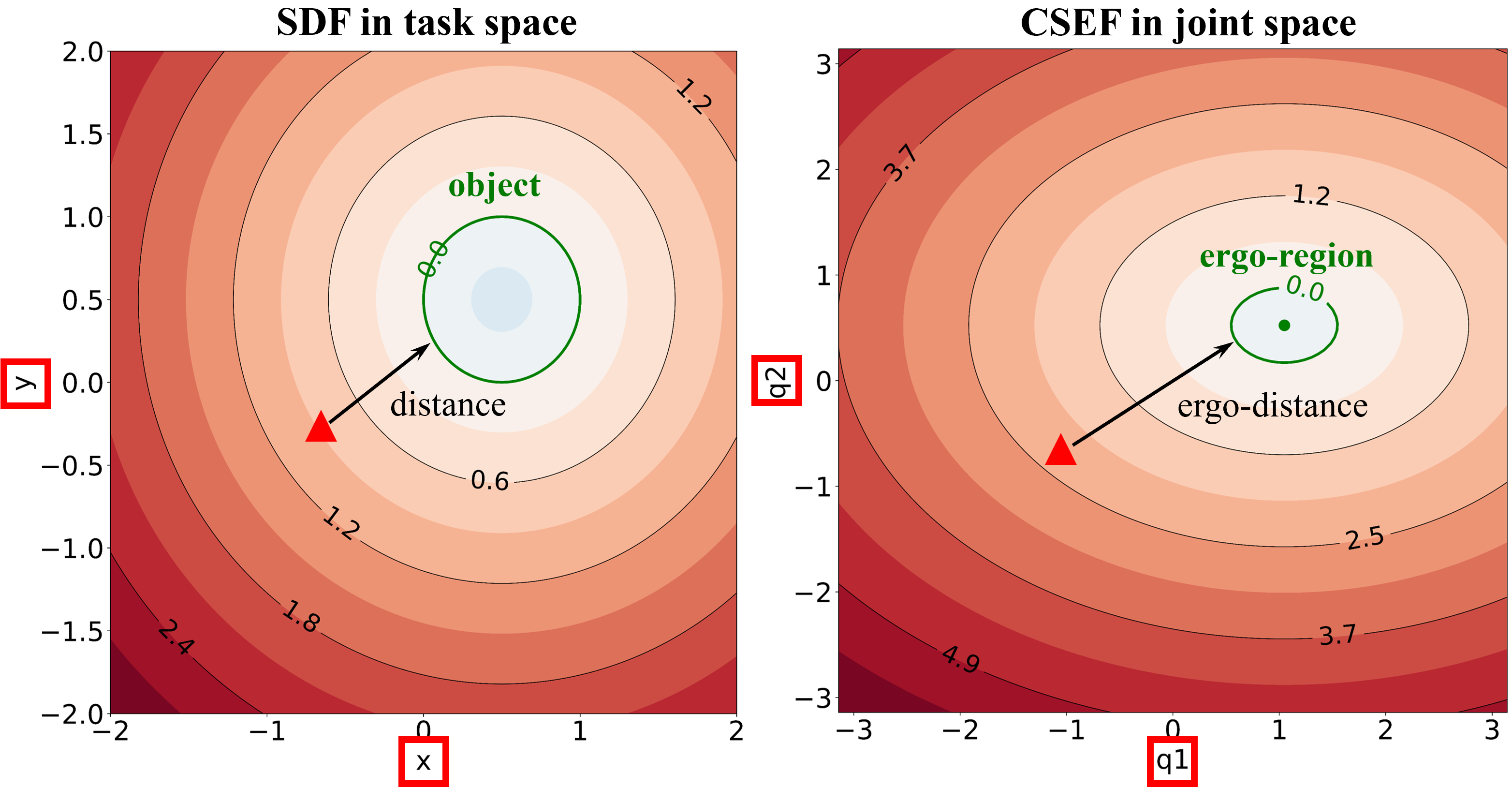} 
\caption{Differences between SDF in task space and CSEF in joint space. The colored level sets depict the distances to the object in SDF, while representing the `ergo-distances' to the defined `ergo-region' in CSEF. \textit{Ergo-distance:} an Euclidean representation based on the ergonomic assessment; \textit{Ergo-region:} a self-designed ergonomic region based on individuals or tasks. }
\label{Fig.csef}
\end{figure}

\subsection{Problem Formulation}

Inspired by the concept of encoding SDF with joint configuration \cite{li2024configuration, koptev2022neural}, CSEF focuses on ergonomic assessment rather than geometric distances from the perspective of humans (Fig. \ref{Fig.csef}). Define $h(\boldsymbol{q})$ as a human posture with joint configuration $\boldsymbol{q} \in \mathbb{R}^n$, where $n$ is the number of degrees of freedom (DoF) in the human kinematic model. CSEF $f_e(\boldsymbol{q})$ is defined as a Euclidean distance function that measures the ergonomic quality of the given joint configuration:

\begin{equation}\label{eq1}
f_e(\boldsymbol{q}) = ||\boldsymbol{q} - \boldsymbol{q}^{opt}||,
\end{equation}
where $\boldsymbol{q}^{opt}$ denotes the optimal ergonomic configuration. In practice, $\boldsymbol{q}^{opt}$ can be specified offline from a chosen ergonomic assessor, or computed by optimization that includes task constraints such as manipulability, tool orientation, and bimanual coordination~\cite{li2025integrating}.

Unlike conventional ergonomic assessment methods that produce discrete scores, CSEF provides a continuous and differentiable representation of ergonomic quality across the entire joint configuration space. This property enables gradient-based optimization and planning for pHRI tasks. Similar to SDFs, CSEF possesses several important properties that make it suitable for ergonomics-aware motion planning. CSEF is differentiable almost everywhere in the joint configuration space, which enables the application of gradient-based optimization methods. The gradient of the CSEF, $\nabla f_e(\boldsymbol{q})$, points in the direction of becoming poor ergonomics, which means that $-\nabla f_e(\boldsymbol{q})$ indicates the direction toward more ergonomically favorable configurations.

The zero-level set of CSEF, ${\boldsymbol{q} \mid f_e(\boldsymbol{q}) = 0}$, represents the optimal posture and serves as a clear target for maintaining ergonomic constraints. Its level sets quantify ergonomic degree. Configurations with similar CSEF values have comparable ergonomic quality, which enables trade-offs among multiple objectives during planning.

For ergonomic optimization, the CSEF can be directly used in gradient projection methods:

\begin{equation}\label{eq2}
\boldsymbol{q}_{t+1} = \boldsymbol{q}_t - a \nabla f_e(\boldsymbol{q}_t),
\end{equation}
where $a$ is a step size parameter. This projection moves the joint configuration toward more comfortable postures. 

\subsection{Computation of CSEF}\label{SEF_compute}

The computation of CSEF involves selecting an appropriate ergonomic assessment model and adding an ergonomic region to calculate the final CSEF value.

\subsubsection{Ergonomic Assessment Model}
Various ergonomic assessment methods exist for evaluating human postures, including RULA, REBA, and OWAS. In this paper, we adapt the RULA model to get the optimal configuration $\boldsymbol{q}^{opt}$.

While these conventional methods typically provide discrete scores, we further modify them as a continuous function of the joint angles and combine them using a weighted sum approach, which can adjust weights based on the specific application context or individual differences \cite{yazdani2022dula}. Hence, $f_e(\boldsymbol{q})$ can be reformulated as a weighted Euclidean distance function based on (\ref{eq1}):
\begin{equation}
f_e(\boldsymbol{q}) = \|\boldsymbol{w}(\boldsymbol{q} - \boldsymbol{q}^{opt})\|,
\end{equation}
where $\boldsymbol{w}$ is a positive diagonal matrix of joint weights that reflect the relative importance of each joint's contribution to overall ergonomics.

\subsubsection{Ergonomic Region}

As mentioned above, when given a human joint configuration $\boldsymbol{q}$, we compute the CSEF value $f_e(\boldsymbol{q})$ as a weighted distance from the optimal configuration. However, the optimal configuration $\boldsymbol{q}^{opt}$ may vary depending on the task context and individual preferences. To account for this variability, we further define a task-specific ergonomic region rather than a single optimal joint set.  Therefore, the CSEF function can be extended as:

\begin{equation}
f_e(\boldsymbol{q}) = \min_{\boldsymbol{q}' \in \Omega} \|\boldsymbol{w}(\boldsymbol{q} - \boldsymbol{q}')\|,
\end{equation}
where $\Omega$ denotes the envelope set bounding the self-designed ergonomically optimal region for the task, and $\boldsymbol{q}'$ is any point set on this envelope.

Then, the gradient of the CSEF with respect to the joint configuration can be computed as:

\begin{equation}\label{gradient}
\nabla f_e(\boldsymbol{q}) = \frac{\boldsymbol{w}^2(\boldsymbol{q} - \boldsymbol{q}'_{\text{min}})}{\|\boldsymbol{w}(\boldsymbol{q} - \boldsymbol{q}'_{\text{min}})\|},
\end{equation}
where $\boldsymbol{q}'_{\text{min}}$ is the closest configuration in the optimal envelope set $\Omega$ to the current configuration $\boldsymbol{q}$. $\boldsymbol{w}^2$ denotes $\boldsymbol{w}^T\boldsymbol{w}$ for diagonal $\boldsymbol{w}$.

\subsection{Task Space Ergonomic Fields (TSEF)}
To complement CSEF in joint space, we project it through forward kinematics to obtain a task space ergonomic field (TSEF). TSEF preserves the ergonomic structure of CSEF while expressing it at end-effector poses, enabling ergonomics-aware reasoning and planning directly in task space. This facilitates fusion with trajectory distance fields, contact and alignment constraints, and other task-space objectives when needed, while the subsequent applications focus primarily on CSEF in configuration space.

Let $\boldsymbol{p} = \mathrm{FK}(\boldsymbol{q}) + \boldsymbol{p}_0 \in \mathbb{R}^m$ be the human end-effector pose given by the human kinematic model, where $\boldsymbol{p}_0$ denotes the pose of the human model’s origin in task space, i.e., the pose obtained from the torso or the scapula. The CSEF $f_e(\boldsymbol{q})$ measures ergonomic cost within the admissible joint set $\Omega_0$. As defined in \cite{merikh2024ergonomically}, for any task space point $\boldsymbol{p}$ intended to be realized by the human end effector, the TSEF $F_e$ is defined as the minimum ergonomic cost among all feasible configurations $\mathcal{Q}(\boldsymbol{p})$ that achieve $\boldsymbol{p}$:
\begin{equation}
\begin{aligned}
&F_e(\boldsymbol{p}) = \min_{\boldsymbol{q} \in \mathcal{Q}(\boldsymbol{p})} f_e(\boldsymbol{q}), \\
&\mathcal{Q}(\boldsymbol{p}) = \bigr\{\boldsymbol{q}\in\Omega_0 \ \big| \ \boldsymbol{q}\in \mathrm{FK}^{-1}(\boldsymbol{p}-\boldsymbol{p}_0) \bigr\} ,
\end{aligned}
\end{equation}
If $\boldsymbol{p}$ is unreachable, $F_e(\boldsymbol{p})$ is assigned a large penalty value and excluded from the feasible planning domain. Therefore, for each $\boldsymbol{p}$, an ergonomic projection $\boldsymbol{q}^\ast(\boldsymbol{p})$ is obtained as a solution of the constrained optimization problem:
\begin{equation}
\boldsymbol{q}^\ast(\boldsymbol{p}) \in \arg\min_{\boldsymbol{q} \in \mathcal{Q}(\boldsymbol{p})} f_e(\boldsymbol{q}),
\end{equation}
Then the TSEF can be defined as:
\begin{equation}\label{F_e}
F_e(\boldsymbol{p})
= f_e\bigl(\boldsymbol{q}^\ast(\boldsymbol{p})\bigr),
\end{equation}
Assuming local uniqueness and differentiability of the selected branch and a full-row-rank Jacobian $\boldsymbol{J}(\boldsymbol{q}^\star)=\frac{\partial \mathrm{FK}}{\partial \boldsymbol{q}}\rvert_{\boldsymbol{q}^\star}$, we have:
\begin{equation}\label{gradient_sef}
\nabla F_e(\boldsymbol{p})
= \bigl(\nabla f_e(\boldsymbol{q}^\star)\bigr)^{T}
\frac{\partial \boldsymbol{q}^\star}{\partial \boldsymbol{p}},
\end{equation}
Applying forward kinematics to \eqref{gradient_sef}, we can get $\frac{\partial \boldsymbol{q}^\star}{\partial \boldsymbol{p}} = \boldsymbol{J}^{\#}(\boldsymbol{q}^\star)$ with $\boldsymbol{J}^{\#}$ the right pseudoinverse. Hence, we can calculate the  gradient as follows:
\begin{equation}
\nabla F_e(\boldsymbol{p})
= \bigl(\nabla f_e(\boldsymbol{q}^\star)\bigr)^{T}\boldsymbol{J}^{\#},
\end{equation}
For numerical robustness near singularities, employ the damped pseudoinverse
$\boldsymbol{J}^{\#}_\lambda
= \boldsymbol{J}^T\bigl(\boldsymbol{J}\boldsymbol{J}^T+\lambda^2 \boldsymbol{I}\bigr)^{-1}
$, the corresponding gradient can be written as:
\begin{equation}
\nabla F_e(\boldsymbol{p})
= \bigl(\nabla f_e(\boldsymbol{q}^\star)\bigr)^{T}\boldsymbol{J}^{\#}_\lambda.
\end{equation}

TSEF is not unique because inverse kinematics yields multiple joint solutions. One end-effector pose can map to several configurations with different ergonomic costs, and this effect increases with redundancy. Prior task-space methods model posture distributions but do not fully capture this multi-solution structure~\cite{figueredo2020human,merikh2024ergonomically}. In our applications, we plan in CSEF where ergonomics is uniquely defined in configuration space and use TSEF only when task-space coupling or fusion with task-space objectives is needed.

\section{Application of CSEF}\label{CSEF_application}
\begin{figure}[!t]
\centering
\includegraphics[width=0.98\linewidth]{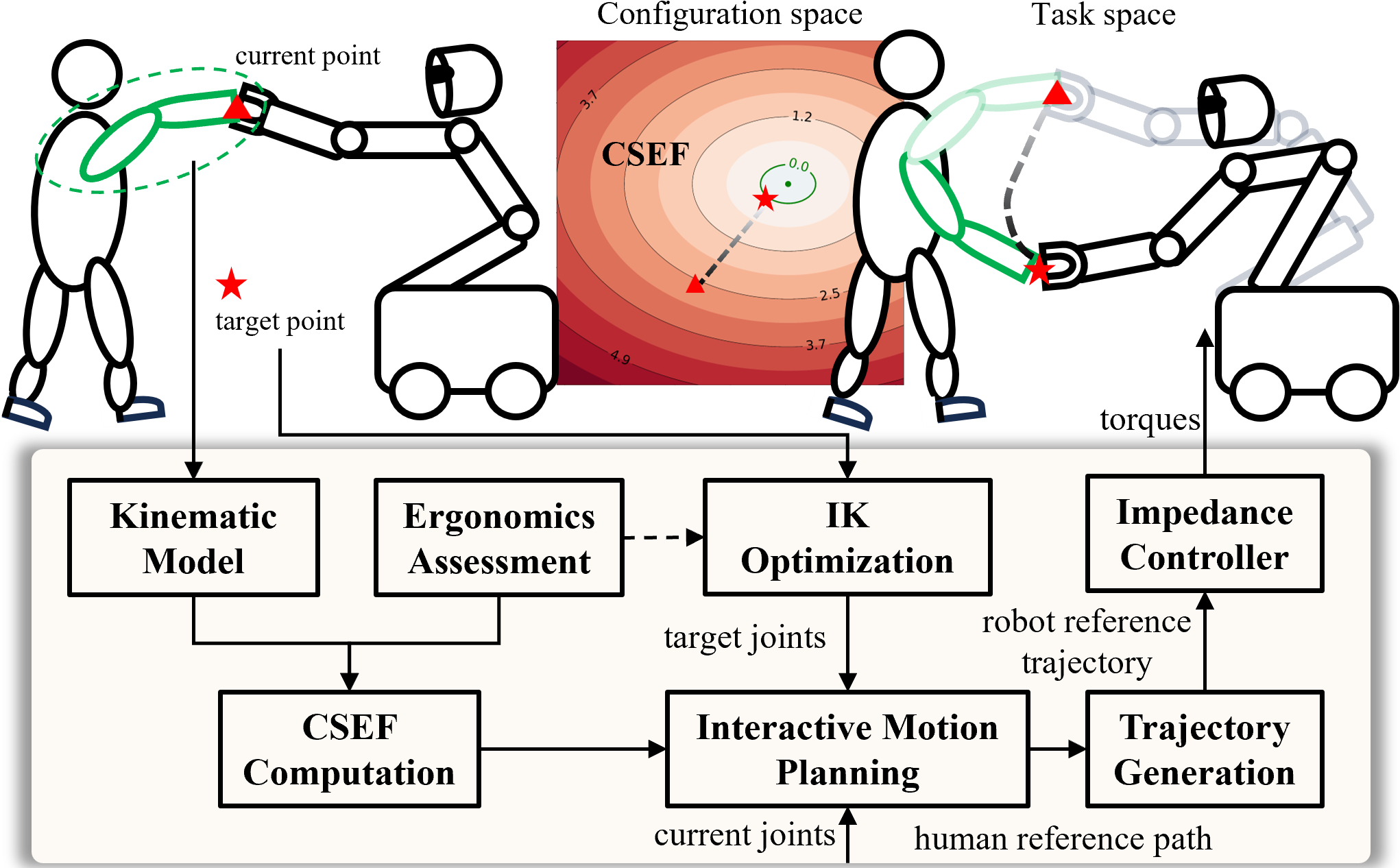} 
\caption{Framework of the CSEF-based interactive motion planning for HRC. \textit{Offline processing:} CSEF can be offline computed based on the human kinematic model and the ergonomic assessment model; \textit{Online execution:} during interaction, the target joint set of the human model is optimized with an inverse kinematics problem; Then the gradient based motion planning can be performed with the CSEF, current joint set, and target joint set; Finally, the trajectory will be executed by an impedance controller of the robot.}
\label{Fig.framework}
\end{figure}

We then demonstrate how CSEF enables ergonomics-aware interactive motion planning in pHRI/HRC across representative collaboration scenarios, as summarized in Fig. \ref{Fig.framework}. 
\subsection{Unimanual Interaction}
Given the current human posture $h(\boldsymbol{q})$ and a point set $\boldsymbol{p}$ in the workspace, the robot should guide the human end effector from the current point to the final point in an ergonomic manner. CSEF evaluates posture ergonomics and provides a continuous field that guides robot motion planning.

\textbf{Case 1 (no target point):} In this case, we directly apply the gradient function (\ref{gradient}) for motion planning.

\textbf{Case 2 (with a defined target point):} Since the target positions are always in the task space, the inverse kinematics problem raises to find the optimal target posture in the configuration space. This problem can be regarded as an optimization problem with the following function:
\begin{equation}
\begin{aligned}
\min_{\boldsymbol{q}' \in \Omega} \| & \boldsymbol{w} (\boldsymbol{q} - \boldsymbol{q}')\| , \\
\text{s.t.} \quad &\boldsymbol{p} = \text{FK}(\boldsymbol{q}), \\
&\boldsymbol{q} \in \Omega_0, \\
\end{aligned}
\end{equation}
where $\boldsymbol{p} = \text{FK}(\boldsymbol{q})$ denotes the forward kinematics of human model, $\Omega_0$ represents the physical boundary range of the human joints. After the target joint set $\boldsymbol{q}^{d}$  is optimized, the motion direction is a weighted combination of a direction vector pointing toward the goal and the negative gradient of the CSEF. The desired motion of the human in the task space can be calculated as the reference path $\boldsymbol{p}_H^{d}$.

\subsection{Bimanual Collaboration}
In real applications, humans and robots may coordinate both end effectors for tasks such as cooperative transport, assembly, or two-handed tool use. These tasks require stable relative poses between the end effectors while keeping human joint configurations ergonomic throughout the motion.

Let the human posture be $h(\boldsymbol{q})$ with two kinematic chains $h_l(\boldsymbol{q}_l)$ and $h_r(\boldsymbol{q}_r)$, whose task-space end effectors are $\boldsymbol{p}_l$ and $\boldsymbol{p}_r$ with desired targets $\boldsymbol{p}^d_l$ and $\boldsymbol{p}^d_r$. Typical bimanual tasks add task-space coupling constraints such as relative pose preservation and bimanual symmetry.

Accordingly, the planning objective is to guide both end-effectors from their current poses toward task-compatible targets while minimizing the CSEF-based ergonomic cost and respecting bimanual constraints. A representative formulation can be given as:
\begin{equation}
\begin{aligned}
&\min_{\boldsymbol{q}' \in \Omega} || \boldsymbol{w}(\boldsymbol{q} - \boldsymbol{q}') || \\
\text{s.t.}\quad
& \boldsymbol{p}_l = \text{FK}_l(\boldsymbol{q}_l),\quad \boldsymbol{p}_r = \text{FK}_r(\boldsymbol{q}_r), \\
&| \ ||\boldsymbol{p}_r - \boldsymbol{p}_l|| - d_{\text{task}} \ | < \epsilon, \\
& \boldsymbol{q} \in \Omega_0, \quad \mathcal{C}(\boldsymbol{q}) \le 0,
\end{aligned}
\end{equation}
where $\text{FK}_l$ and $\text{FK}_r$ map the joint configuration to the left/right end-effector frames, $d_{\text{task}}$ is the relative contact distance. The weighting matrix $\boldsymbol{w}$ and $\boldsymbol{q}'$ follow the same definitions as in the unimanual case, capturing ergonomic preferences in joint space. $\mathcal{C}$ represents additional feasibility conditions to avoid collisions or ensure manipulability/affordance. 

After obtaining the target joint configuration $\boldsymbol{q}_l^{d}$ and $\boldsymbol{q}_r^{d}$, we perform motion planning based on the CSEF representation in the joint space by applying gradient-based motion optimization approaches such as quadratic programming (QP). Finally, the desired motion of the human in the joint space can be transformed into the task space by forward kinematics, where the robot will interactively guide the human to follow the reference paths $\boldsymbol{p}_{H\_l}^{d}$ and $\boldsymbol{p}_{H\_r}^{d}$. 

\subsection{Trajectory Generation and Robot Controller}
\begin{figure}[!t]
\centering
\includegraphics[width=0.95\linewidth]{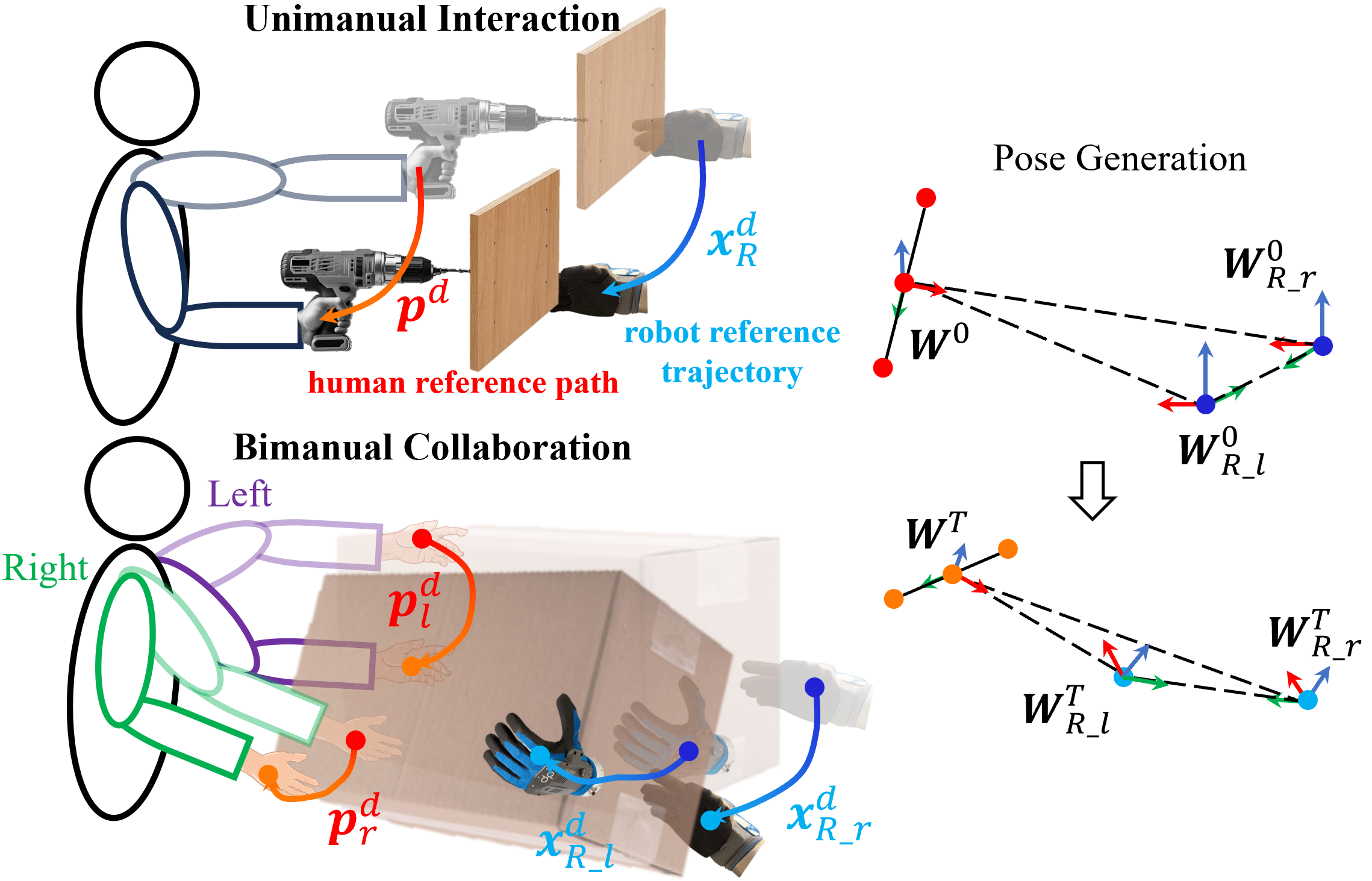} 
\caption{Illustration of robot reference trajectory generation based on the human reference path in unimanual interaction and bimanual collaboration.}
\label{Fig.pose_generation}
\vspace{-4mm}
\end{figure}

We then propose a robot impedance controller that requires an end-effector reference $\boldsymbol{x}_R^d$ for motion execution to gently guide the human to complete posture adjustment. Hence, we map $\boldsymbol{p}^d$ to $\boldsymbol{x}_R^d$ in two common cases shown in Fig. \ref{Fig.pose_generation}.

\textbf{Direct Unimanual Interaction:} The robot end effector is rigidly coupled to the human endpoint with a constant relative pose. Let $T_R^H$ be the fixed transform from the human frame to the robot frame at the contact. Then the robot reference position is generated incrementally from the human reference:
\begin{equation}
\boldsymbol{p}_R^d =  T_R^H \boldsymbol{p}_H^d,
\end{equation}
The orientation of the robot end effector can be adjusted according to task requirements or remain unchanged.

\textbf{Bimanual Object Manipulation:} The human and robot grasp a shared object, so the robot reference trajectories follow from the contact constraints. As shown in Fig. \ref{Fig.pose_generation}, at the initial moment, we take the positions of the human left and right end effectors and define their midpoint as the origin $\boldsymbol{W}^0$. The $Y$-axis points from the left to the right wrist. Using the plane defined by $\boldsymbol{W}^0$ and the two robot end-effector poses ($\boldsymbol{W}_{R\_l}^0$ and $\boldsymbol{W}_{R\_r}^0$), we select an $X$-axis that lies in this plane and is orthogonal to the $Y$-axis. The $Z$-axis is the normal of this plane, which completes a right-handed frame. With the initial poses of the robot end effectors $\boldsymbol{x}_{R\_l}^0$ and $\boldsymbol{x}_{R\_r}^0$, the corresponding vectors from the initial object pose to each end effector pose are represented as $\boldsymbol{v}_{R\_l}$ and $\boldsymbol{v}_{R\_r}$.

The human reference paths are generated based on the CSEF, which are defined as $\boldsymbol{p}_{H\_l}^{d}$ and $\boldsymbol{p}_{H\_r}^{d}$. We then calculate the midpoint at any timestep as $\boldsymbol{W}^T$. The rotation matrix between $\boldsymbol{W}^0$ and $\boldsymbol{W}^T$ can be given as $\boldsymbol{R}_{0}^T$. Hence, the reference positions of the robot end effector are calculated as $\boldsymbol{p}_{R\_l}^T$ and $\boldsymbol{p}_{R\_r}^T$. Similarly, the reference orientations are generated by multiplying the initial rotation with the rotation matrix $\boldsymbol{R}_{0}^T$. Finally, the reference poses at $T$ are given as $\boldsymbol{x}_{R\_l}^T$ and $\boldsymbol{x}_{R\_r}^T$.

In HRC, the motion of the robot end effector arises from the combined influence of the robot control input, denoted as $\boldsymbol{F}^r$, and the external force $\boldsymbol{F}^e$ applied by the human. Assuming $\boldsymbol{F}^r$ is generated by a virtual spring-damper system, the interaction dynamics can be described as:

\begin{equation}\label{interaction_model}
\boldsymbol{M^d}\boldsymbol{\ddot{x}}_{R} = \boldsymbol{K}^d (\boldsymbol{x}_R^d - \boldsymbol{x}_R) + \bm{D}^d (\boldsymbol{\dot{x}}_R^d - \boldsymbol{\dot{x}}_R) + \boldsymbol{F}^e,
\end{equation}
where $\boldsymbol{x}_R$ and $\boldsymbol{x}_R^d$ represent the current and reference poses of the end effector, respectively, while $\boldsymbol{M}^d$, $\boldsymbol{K}^d$, and $\boldsymbol{D}^d$ are the inertia, stiffness, and damping matrices. 

The robot joint-space dynamics are governed by:
\begin{equation}\label{robot_dynamics}
\boldsymbol{M}(\boldsymbol{q}_R)\boldsymbol{\ddot{q}}_R + \boldsymbol{C}(\boldsymbol{q}_R, \boldsymbol{\dot{q}}_R)\boldsymbol{\dot{q}}_R + \boldsymbol{g}(\boldsymbol{q}_R)
= \boldsymbol{\tau} + \boldsymbol{\tau}^e,
\end{equation}
where $\bm{q}_R$ represents the joint angles, with $\boldsymbol{M}(\boldsymbol{q}_R)$ being the joint-space inertia matrix, $\boldsymbol{C}(\boldsymbol{q}_R, \boldsymbol{\dot{q}}_R)$ the Coriolis and centrifugal matrix, and $\boldsymbol{g}(\boldsymbol{q}_R)$ the gravity vector. The control torque $\boldsymbol{\tau}$ and the external torque $\boldsymbol{\tau}^e = \bm{J}_R^T \boldsymbol{F}^e$ act in joint space, where $\bm{J}_R$ denotes the robot Jacobian matrix.

By integrating the interaction model in (\ref{interaction_model}) with the robot dynamics in (\ref{robot_dynamics}), the joint torques $\boldsymbol{\tau}$ can be computed. These torques depend on the reference trajectory generated by CSEF-based planning and the defined impedance and damping parameters.

\section{Experiments}
To verify the efficacy of CSEF in interactive motion planning, we initially explored qualitative results through a 2-DoF example in simulation, then progressed to a more realistic 4-DoF human upper limb model in real-world experiments by physical robot guidance, unimanual collaborative drilling, and bimanual collaborative carrying tasks.

\begin{figure*}[!t]
\centering
\includegraphics[width=0.95\linewidth]{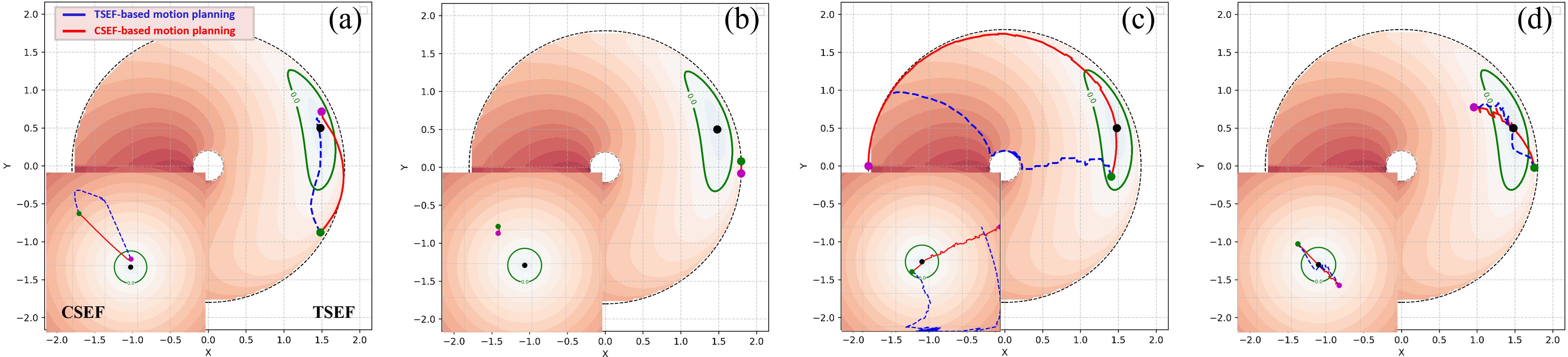} 
\caption{Simulation results of comparison between CSEF-based and TSEF-based motion planning in 2D example. The green and purple solid circles represent the start and end points, respectively. The red solid lines and blue dashed lines are the planned trajectories based on CSEF and TSEF, respectively. Four cases are given: (a) from high-ergo point to low-ergo point; (b) near singularity; (c) from low-ergo point to high-ergo point; (d) crossing the `ergo-region'.}
\label{Fig.simulation_2D}
\vspace{-4mm}
\end{figure*} 

\begin{table*}[ht]
\centering
\setlength{\tabcolsep}{3.5pt}
\footnotesize
\caption{Comparison between a CSEF-based planner and a TSEF-based planner}
\label{tab:comparison}
\begin{tabular}{@{}lcccccccc@{}}
\toprule
Metrics & \multicolumn{2}{c}{Case 1} & \multicolumn{2}{c}{Case 2} & \multicolumn{2}{c}{Case 3} & \multicolumn{2}{c}{Case 4} \\
\cmidrule(lr){2-3} \cmidrule(lr){4-5} \cmidrule(lr){6-7} \cmidrule(lr){8-9}
& \textit{CSEF} & \textit{TSEF} & \textit{CSEF} & \textit{TSEF} & \textit{CSEF} & \textit{TSEF} & \textit{CSEF} & \textit{TSEF} \\ 
\midrule
Average CSEF Value & \textbf{0.6754} & 1.9762 & \textbf{0.7667} & NA & \textbf{0.7212} & NA & 0.1420 & \textbf{0.0639} \\
Maximum CSEF Value & \textbf{1.7862} & 2.3877 & \textbf{0.8903} & NA & \textbf{2.0784} & NA & 0.3485 & 0.3485 \\
Cartesian Path Length (m) & 1.7386 & 1.7407 & 0.1601 & NA & 5.5767 & NA & 1.8729 & 1.6774 \\
Joint Space Path Length (rad) & 2.2091 & 3.8746 & 0.2875 & NA & 3.7462 & NA & 2.0926 & 2.5266 \\
Computation Time (s) & \textbf{0.0015} & 14.9545 & \textbf{0.0003} & 366.0760 & \textbf{0.0025} & 414.7832 & \textbf{0.0014} & 14.9510 \\
\midrule
\multicolumn{9}{l}{\footnotesize \textasteriskcentered\ \textbf{Success rate} (100 random cases) \quad CSEF: 100\% \quad TSEF: 62\%} \quad NA: exceeded the maximum iterations \\
\bottomrule
\end{tabular}
\vspace{-3mm}
\end{table*}

\subsection{Simulation with 2-DoF Planar Model}

In simulation, we implement a 2-DoF planar model representing a simplified human arm that interacts with a robotic manipulator in a 2D workspace. This model consists of two links with lengths $l_1 = 1.0 \ m$ and $l_2 = 0.8 \ m$, representing the upper arm and forearm, respectively, with joint limits $q_1, q_2 \in [-\pi, \pi]$. The ergonomic optimal configuration for this simplified model is set at $q_1^{opt} = \pi/4$ and $q_2^{opt} = -\pi/3$. The ergonomic threshold is set to $\epsilon = 0.5$, with joint weights $\boldsymbol{w} = [1.0, 1.0]$ giving equal importance to both joint configurations. The CSEF in joint space is computed using the methodology outlined in Section \ref{SEF_compute}. 

We introduce the gradient-based motion planner used. Considering \textbf{Case 2} of Section \ref{CSEF_application}, we apply a potential field with attraction to the goal and repulsion from high CSEF values. The motion direction is a weighted sum of a goal-directed vector, the negative CSEF gradient, and a small random perturbation to escape local minima. We then construct TSEF via forward kinematics for task-space planning as shown in Fig. \ref{Fig.simulation_2D}. For comparison, we use the ergonomically optimized planner of \cite{merikh2024ergonomically} and perform motion planning on the TSEF.

The performance of CSEF and TSEF motion planning approaches is compared based on several key metrics, including average CSEF value, maximum CSEF value, Cartesian path length, joint-space path length, and computation time. The quantitative results are summarized in Table~\ref{tab:comparison}, and the trajectory planning results are visualized in Fig.~\ref{Fig.simulation_2D}.

The success rate, evaluated over 100 random cases, demonstrates the robustness of joint-space planning. our CSEF method achieves a success rate of $100\%$, compared to $62\%$ for TSEF-based planning. This disparity can be attributed to the singularity problem, as shown in Fig.~\ref{Fig.simulation_2D}(b), or the higher sensitivity of task-space planning to local minima in CSEF, as evidenced in Fig.~\ref{Fig.simulation_2D}(c). Specifically, the TSEF-based planning can not even find the solution in Case 2 while failing to reach the target point set in Case 3. Besides, the trajectories generated by TSEF-based planning often encounter regions of high CSEF (shown in Fig.~\ref{Fig.simulation_2D}(a)), which is due to the multiple solutions in inverse kinematics problems.

Across cases, the CSEF-based planner yields lower average CSEF values than the TSEF-based planner. For example, in Case 1 the average value with CSEF is 0.6754 versus 1.9762 with TSEF. In Case 4, where the path crosses the “ergo-region,” the TSEF-based method performs slightly better. Similar trends hold for maximum CSEF values, indicating that configuration-space planning produces trajectories more stable under the CSEF metric, which is important for physical ergonomics. CSEF-based planning also generates shorter paths in configuration space. In computation time, Table~\ref{tab:comparison} shows that CSEF-based planning is more efficient. Trajectory visualizations in Fig.~\ref{Fig.simulation_2D} further illustrate these differences.

\begin{figure}[!t]
\centering
\includegraphics[width=0.95\linewidth]{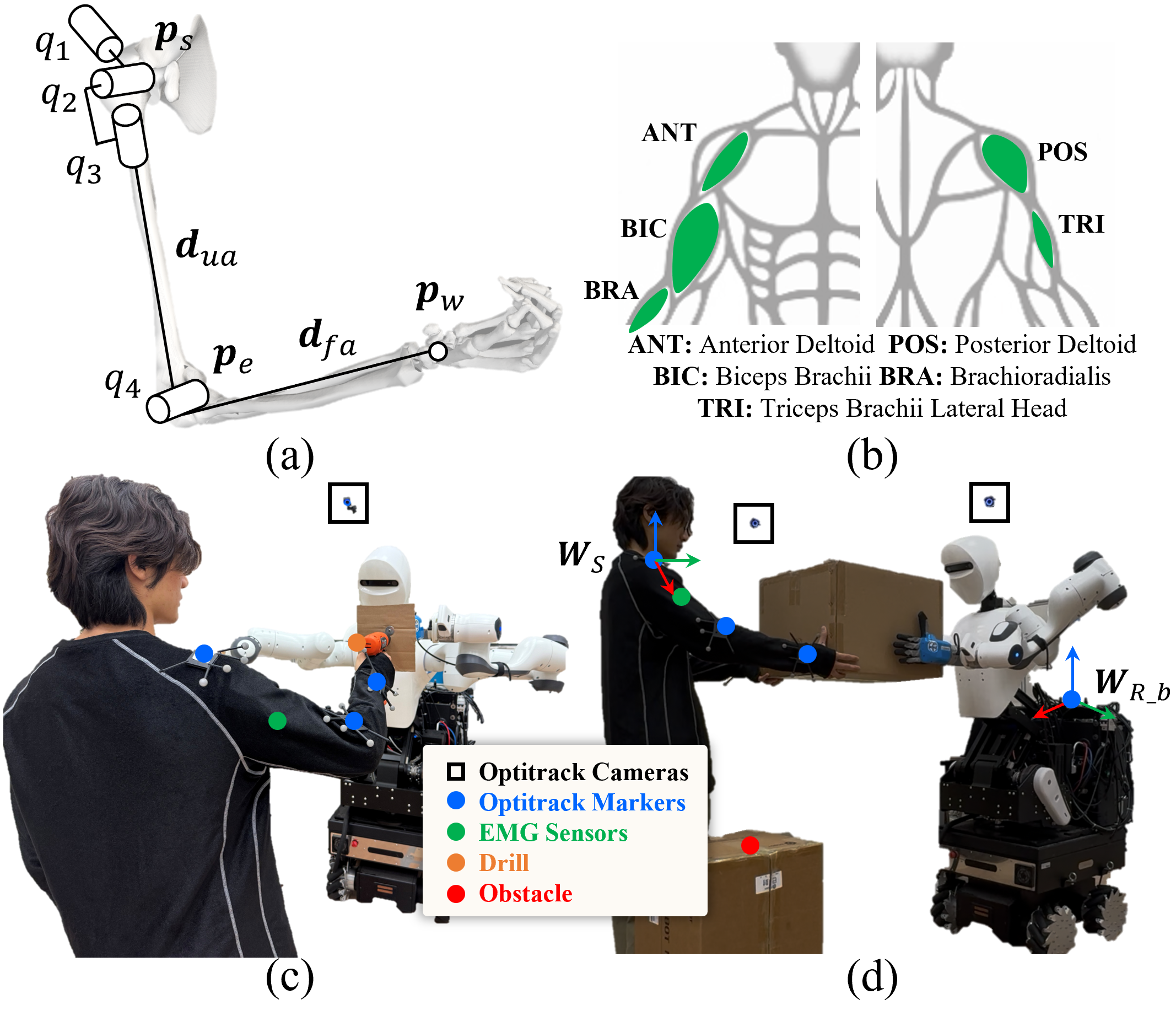} 
\caption{Experimental setup. (a): the simplified human right upper limb skeleton model with three joints on the shoulder and one joint on the elbow; (b): Target muscles monitored for further validation; (c-d): illustration of the human-robot collaborative drilling/carrying tasks with the black rectangles/blue solid circles represent the motion capture cameras/markers, green solid circles be the EMG sensors, orange solid circle be the drill, and red solid circle be the obstacle.}
\label{Fig.exp} 
\vspace{-4mm}
\end{figure}

\subsection{HRC with 4-DoF Human Upper Limb Model}
\subsubsection{Unimanual Human-Robot Interaction}
To validate our approach in more realistic scenarios, a pHRI task is designed based on \textbf{Case 1} in which the robot needs to lead the human from its current point to the ergonomically-optimized point. During interaction, the robot will act as a leader while the human will perform as a follower with their end effector fixed to the other. In this case, a 4-DoF human upper limb skeleton model is applied for human kinematics modeling, which includes three degrees of freedom at the shoulder (abduction/adduction, flexion/extension, and internal/external rotation) and one degree of freedom at the elbow (flexion/extension) (shown in Fig. \ref{Fig.exp}(a)). This model more accurately captures the biomechanical complexity of human arm movements during physical interaction tasks.

The joint limits for this model are set according to anatomical constraints: shoulder abduction/adduction $q_1 \in [-\pi/18, 17\pi/18]$, shoulder flexion/extension $q_2 \in [-\pi/3, 17\pi/18]$, shoulder internal/external rotation $q_3 \in [-\pi/3, \pi/2]$, and elbow flexion/extension $q_4 \in [-\pi/2, \pi/3]$. The optimal ergonomic configuration varies based on the specific task, but generally favors neutral positions with minimal joint strain. In this paper, the optimal ergonomic configuration is set as $\boldsymbol{q}^{opt} = [0, 0, 0, \pi/6]$ while the weight matrix is set as $\boldsymbol{w} = [1.0, 1.0, 1.0, 2.0]$ based on the joint risk levels in the REBA method \cite{hignett2000rapid}.

\begin{figure}[!t]
\centering
\includegraphics[width=1\linewidth]{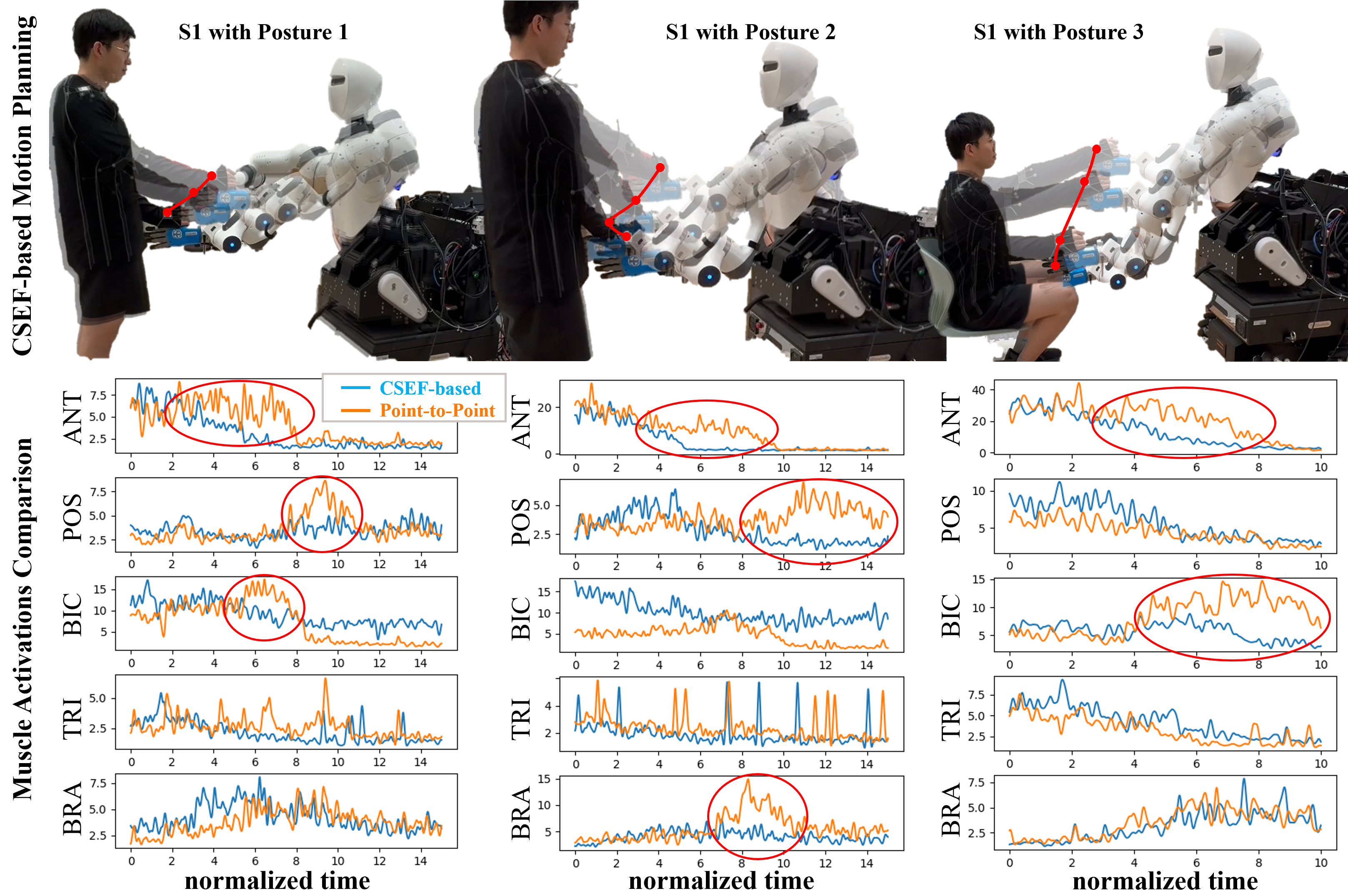} 
\caption{Experimental validation by interactive motion planning based on CSEF with three different initial postures. Comparison of Muscle Activation with the Point-to-Point motion planning method is given.}
\label{Fig.exp_hri}
\vspace{-4mm}
\end{figure}

We conducted real-world experiments with a dual-arm collaborative robot and human subjects. The platform is a mobile manipulator with two torque-controlled 7-DoF Franka Emika Panda arms, controlled via impedance as in \cite{li2024towards}. Subject motion was captured with OptiTrack, and the recorded postures were processed by the CSEF-based planner to compute trajectories. The robot transformed trajectories from the human shoulder frame $\boldsymbol{W}^S$ to the robot base frame $\boldsymbol{W}^{R_b}$ and executed them with impedance control to guide the human hand from the current to the target point. Muscle activation was measured using sEMG to quantify changes from initial to target postures with five chosen muscles (Fig.~\ref{Fig.exp}(b)).

Fig.~\ref{Fig.exp_hri} shows snapshots of robot guidance for three initial postures. The target posture in all trials is the optimal joint configuration. Solid red lines indicate the human hand motion during interaction. To assess ergonomic benefits, we ran comparative trials that used the shortest path in task space. Minimum-jerk trajectories were generated for the robot’s left arm from the initial to the target point in task space.

Muscle activation was also compared during pHRI for CSEF-based and PTP trajectories, with blue for CSEF and orange for PTP (shown in Fig.~\ref{Fig.exp_hri}). The CSEF-based planner yields a consistently decreasing activation profile across most muscles (ANT, POS, BIC, TRI, BRA), whereas PTP shows frequent fluctuations and reactivations. This contrast is most evident in triceps and biceps, where CSEF suppresses the activation spikes typical of conventional paths. A stable decline reduces unnecessary strain, fatigue, and co-contraction, potentially lowering repetitive strain injury risk and offering a more ergonomic strategy for robotic assistance.

\begin{figure*}[!t]
\centering
\includegraphics[width=0.95\linewidth]{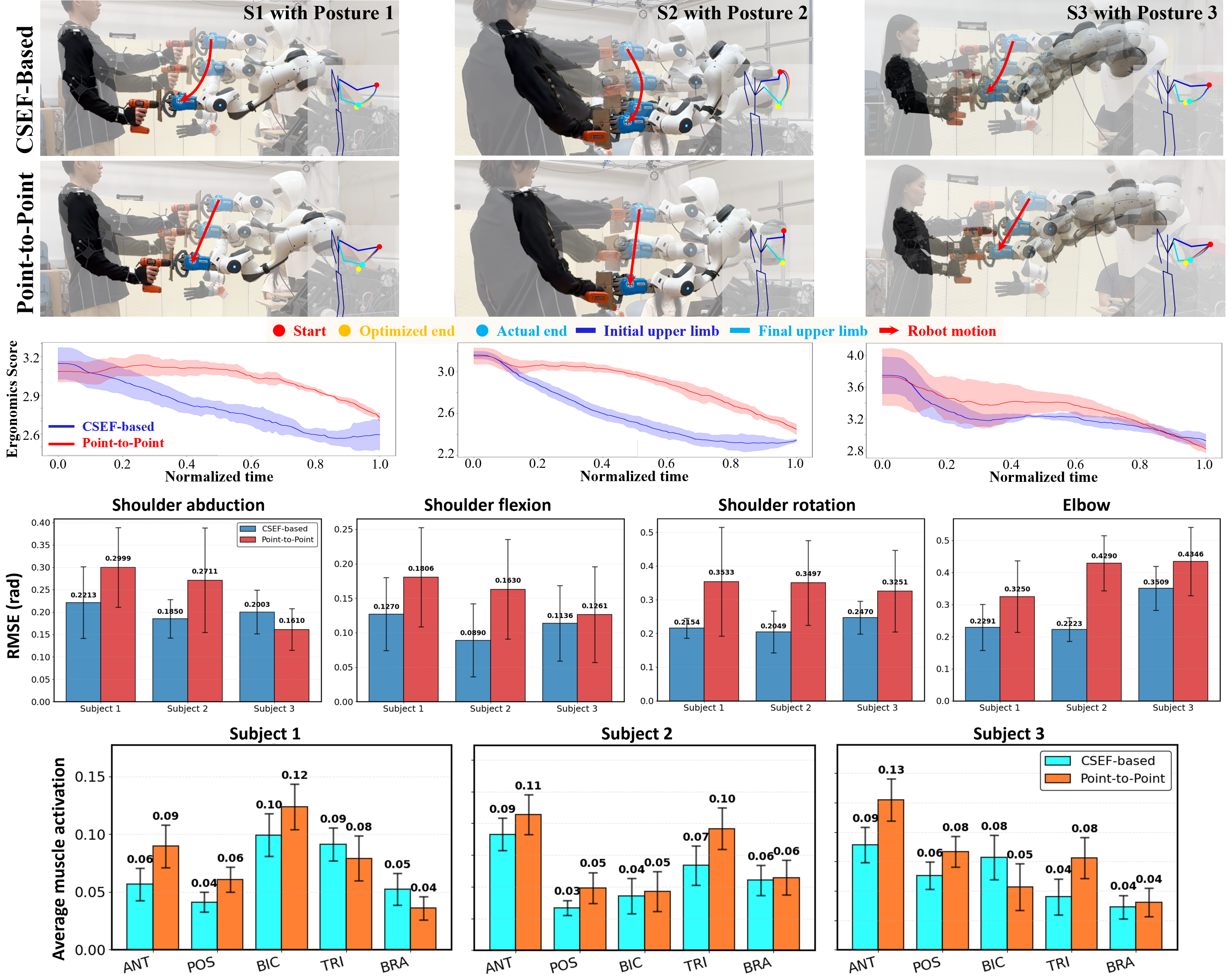} 
\caption{The snapshots of human-robot collaborative drilling tasks with CSEF-based motion planning and Point-to-Point motion planning. The adjusted trajectories of 3 subjects (S1-S3) under 3 initial postures under the guidance of the robot are shown with a human skeleton model. The comparison of the ergonomic score variation, RMSE ($rad$) between the actual joint angle trajectories and the optimized joint angle trajectories, and average muscle activation of five target muscles is given.}
\label{Fig.exp_drilling}
\vspace{-4mm}
\end{figure*}

\subsubsection{Unimanual Collaborative Drilling}
We designed a collaborative drilling task to assess CSEF-based planning, which demands precise coordination and ergonomic guidance. As shown in Fig. \ref{Fig.exp}(c), the robot grasps and stabilizes a workpiece while the human operates a handheld drill at the target location. Each trial starts with the robot placing the workpiece at a random reachable pose in the shared workspace. The human chooses a comfortable initial posture, after which the robot estimates the human joint configuration from motion capture and plans an ergonomically favorable motion using CSEF, respecting joint limits and task feasibility. The robot executes the plan via impedance control to provide compliant guidance toward the desired drilling configuration. Upper-limb joint angles and muscle activation are recorded synchronously with motion capture and sEMG. 3 subjects are tested with 3 distinct initial postures, and each subject performs 3 trials per posture, yielding 27 trials in total.

To evaluate ergonomic guidance, we compare CSEF-based planning with a PTP minimum-jerk task-space planner that ignores ergonomics, using identical initial and target drilling points (Fig. \ref{Fig.exp_drilling}). Ergonomic-score trajectories show that CSEF reduces cost faster from the onset and maintains lower values over a larger portion of the motion, indicating quicker convergence to favorable joint configurations and less time in suboptimal postures. The benefit is most evident for S1 and S2, whose average ergonomic score were reduced by 7.28\% and 10.31\%, respectively. S3 also improves, with a smaller 3.11\% reduction, indicating inter-subject differences in responsiveness, potentially due to joint flexibility or preferred movement strategies.

We further quantify performance by comparing joint-angle tracking error between the CSEF-planned optimal trajectory and the executed human motion, and by analyzing average activation of five muscles (Fig. \ref{Fig.exp_drilling}). Across all four joints, CSEF yields lower mean errors than the PTP baseline for all subjects, with the largest gains at the shoulder where geometric tracking induces less ergonomic, harder-to-follow motions. For S1 and S2 the error gap is pronounced and aligns with their faster ergonomic-score descent, while S3 shows a smaller but consistent improvement, indicating subject-dependent sensitivity to ergonomic shaping.

Muscle activation results similarly favor CSEF. Average activation is reduced for most monitored muscles relative to PTP, with clear decreases in anterior and posterior deltoid and biceps for S1–S2, and consistent reductions for S3. Triceps and brachialis also trend lower. Lower means with smaller dispersions suggest that CSEF reduces overall effort and suppresses transient spikes, yielding a more economical and repeatable neuromuscular response.

\subsubsection{Bimanual Collaborative Carrying}
\begin{figure*}[!t]
\centering
\includegraphics[width=0.92\linewidth]{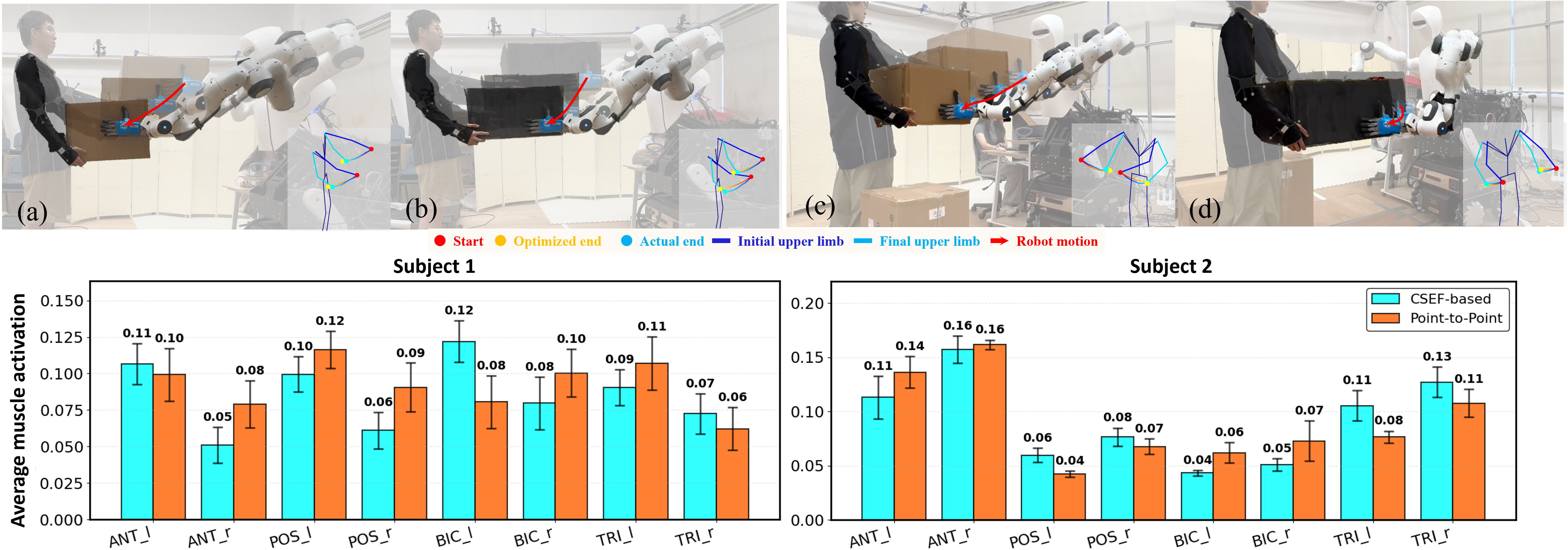} 
\caption{Experimental validation by CSEF-based bimanual motion planning in bimanual human-robot collaborative carrying tasks. (a-b): Subject 1 with two initial postures; (c-d): Subject 2 with two initial postures. The average muscle activation of the target muscles is compared with PTP baseline. }
\label{Fig.exp_carrying} 
\vspace{-4mm}
\end{figure*}

We also tested the framework in a bimanual co-carrying task that demands tight coordination between a human and a dual-arm robot (Fig. \ref{Fig.exp_carrying}). The human and robot jointly grasp a large object while maintaining a task-space constraint on the relative pose between their end effectors. Each human limb is assigned its own CSEF encoding arm-specific ergonomic cost and gradients, and the two CSEFs are optimized jointly with a coupling constraint that preserves a fixed inter-end-effector distance, enabling ergonomics-aware assistance under shared-object geometry.

During collaboration, the human’s upper limb joint kinematics are captured via motion tracking, and the ergonomics of each arm are evaluated in real time using the corresponding CSEF. Given the chosen initial posture, the planner computes a bimanual trajectory that simultaneously improves each limb's ergonomics according to its respective CSEF, satisfies anatomical joint limits, and respects the inter-arm relative-distance constraint throughout the motion. The robot executes the planned motion using impedance control at both arms to maintain compliant physical interaction while stabilizing the object and accommodating human micro-adjustments. As with the drilling study, a PTP task-space planner serves as a baseline, generating minimum-jerk trajectories that meet the same end poses but without explicit ergonomic optimization. Two human subjects participated in the study. Two large objects with three initial postures of subjects were provided to the planner, yielding diverse starting configurations.

The bimanual carrying results in Table \ref{tab:bimanual_erg} indicate consistent ergonomic benefits for the CSEF-based planner over the PTP baseline across both subjects and arms. For S1, CSEF reduces the average ergonomic score for both the left arm (5.34\%) and right arm (5.60\%), with smaller standard deviations suggesting more stable ergonomic quality during execution. A similar trend appears for S2, where the right arm shows a clear improvement (2.80\%), and the left arm is at least comparable while slightly better on average (0.69\%). Overall, CSEF achieves lower ergonomic cost in all cases, with notably reduced variability for S1, indicating that incorporating ergonomic gradients into planning yields both improved average ergonomic scores and steadier performance under the inter-arm coupling constraint.

Bar charts for S1 and S2 in Fig. \ref{Fig.exp_carrying} show a consistent advantage for the CSEF planner across most monitored muscles on both arms. For S1, CSEF lowers activation in anterior and posterior deltoid and triceps bilaterally, with a marked reduction in left biceps where PTP shows the highest load. S2 displays a similar pattern, with CSEF at or below PTP for most muscles, especially posterior deltoid and biceps, while residual differences in right anterior deltoid are small and within variability.

Overall, CSEF shifts effort away from peak-demand groups linked to shoulder elevation and elbow flexion, producing a more balanced neuromuscular profile under constrained co-manipulation. The largest reductions occur in biceps and triceps, which support and stabilize the payload, indicating improved mechanical advantage and fewer corrective activations under the inter-arm distance constraint. These trends align with the lower ergonomic scores, suggesting that ergonomically informed planning improves posture and yields measurable decreases in muscular demand during bimanual carrying.

\begin{table}[t]
\centering
\small
\setlength{\tabcolsep}{5pt}
\caption{Average Ergonomic Score Comparison of Bimanual Collaborative Carrying (mean $\pm$ sd)}
\label{tab:bimanual_erg}
\begin{tabular}{@{}lcccc@{}}
\toprule
\multirow{2}{*}{Subject} & \multirow{2}{*}{Method} & \multicolumn{2}{c}{Ergonomic score} \\
& & Left & Right \\
\midrule
\multirow{2}{*}{S1}  & CSEF-based
& $2.66 \pm 0.13$
& $2.36 \pm 0.04$
 \\
 & Point-to-Point
& $2.81 \pm 0.21$
&$2.50 \pm 0.06$
 \\
 \addlinespace[0.2em]
\hline
\addlinespace[0.2em]
\multirow{2}{*}{S2}  & CSEF-based
& $2.87 \pm 0.14$
& $2.08 \pm 0.07$
 \\
& Point-to-Point
& $2.89 \pm 0.10$
& $2.14 \pm 0.03$
 \\
\bottomrule
\end{tabular}
\vspace{-4mm}
\end{table}

\section{Conclusions}
This work introduced the configuration space ergonomic field as a continuous, differentiable representation of physical ergonomics in joint space and demonstrated its use for interactive, real-time planning in HRC. By framing ergonomics as a field over configurations, the method exposes informative gradients that guide humans toward safer postures while respecting task feasibility and joint limits. We further projected CSEF through forward kinematics to obtain a Task Space Ergonomic Field for end-effector reasoning and fusion with task objectives, enabling a unified treatment of ergonomics and task performance.

The results indicate strong industrial viability. In simulation, CSEF exceeded a task-space ergonomics-based planner in success rate and computation time. On hardware, it delivered faster ergonomic cost reduction, smaller joint tracking errors to optimized references, and lower average muscle activation with fewer transients across unimanual guidance, collaborative drilling, and bimanual co-carrying. The results show that configuration-space ergonomic representations can be seamlessly integrated into standard impedance-controlled stacks and deployed on the shop floor without compromising throughput or safety.

\bibliographystyle{ieeetr}
\bibliography{ref_peo}

@article{da2010risk,
  title={Risk factors for work-related musculoskeletal disorders: a systematic review of recent longitudinal studies},
  author={Da Costa, Bruno R and Vieira, Edgar Ramos},
  journal={American journal of industrial medicine},
  volume={53},
  number={3},
  pages={285--323},
  year={2010},
  publisher={Wiley Online Library}
}

@inproceedings{van2020predicting,
  title={Predicting and optimizing ergonomics in physical human-robot cooperation tasks},
  author={van der Spaa, Linda and Gienger, Michael and Bates, Tamas and Kober, Jens},
  booktitle={2020 IEEE International Conference on Robotics and Automation (ICRA)},
  pages={1799--1805},
  year={2020},
  organization={IEEE}
}

@article{figueredo2020human,
  title={Human comfortability: Integrating ergonomics and muscular-informed metrics for manipulability analysis during human-robot collaboration},
  author={Figueredo, Luis FC and Aguiar, Rafael Castro and Chen, Lipeng and Chakrabarty, Samit and Dogar, Mehmet R and Cohn, Anthony G},
  journal={IEEE Robotics and Automation Letters},
  volume={6},
  number={2},
  pages={351--358},
  year={2020},
  publisher={IEEE}
}

@article{koptev2022neural,
  title={Neural joint space implicit signed distance functions for reactive robot manipulator control},
  author={Koptev, Mikhail and Figueroa, Nadia and Billard, Aude},
  journal={IEEE Robotics and Automation Letters},
  volume={8},
  number={2},
  pages={480--487},
  year={2022},
  publisher={IEEE}
}

@article{hignett2000rapid,
  title={Rapid entire body assessment (REBA)},
  author={Hignett, Sue and McAtamney, Lynn},
  journal={Applied ergonomics},
  volume={31},
  number={2},
  pages={201--205},
  year={2000},
  publisher={Elsevier}
}

@inproceedings{li2024towards,
  title={Towards Robo-Coach: Robot Interactive Stiffness/Position Adaptation for Human Strength and Conditioning Training},
  author={Li, Chenzui and Wu, Xi and Teng, Tao and Calinon, Sylvain and Chen, Fei},
  booktitle={2024 IEEE International Conference on Robotics and Automation (ICRA)},
  pages={860--866},
  year={2024},
  organization={IEEE}
}

@article{lorenzini2023ergonomic,
  title={Ergonomic human-robot collaboration in industry: A review},
  author={Lorenzini, Marta and Lagomarsino, Marta and Fortini, Luca and Gholami, Soheil and Ajoudani, Arash},
  journal={Frontiers in Robotics and AI},
  volume={9},
  pages={813907},
  year={2023},
  publisher={Frontiers Media SA}
}

@inproceedings{busch2017postural,
  title={Postural optimization for an ergonomic human-robot interaction},
  author={Busch, Baptiste and Maeda, Guilherme and Mollard, Yoan and Demangeat, Marie and Lopes, Manuel},
  booktitle={2017 IEEE/RSJ International Conference on Intelligent Robots and Systems (IROS)},
  pages={2778--2785},
  year={2017},
  organization={IEEE}
}

@article{kim2017anticipatory,
  title={Anticipatory robot assistance for the prevention of human static joint overloading in human--robot collaboration},
  author={Kim, Wansoo and Lee, Jinoh and Peternel, Luka and Tsagarakis, Nikos and Ajoudani, Arash},
  journal={IEEE robotics and automation letters},
  volume={3},
  number={1},
  pages={68--75},
  year={2017},
  publisher={IEEE}
}

@article{mcatamney1993rula,
  title={RULA: a survey method for the investigation of work-related upper limb disorders},
  author={McAtamney, Lynn and Corlett, E Nigel},
  journal={Applied ergonomics},
  volume={24},
  number={2},
  pages={91--99},
  year={1993},
  publisher={Elsevier}
}

@inproceedings{yazdani2022dula,
  title={DULA and DEBA: Differentiable ergonomic risk models for postural assessment and optimization in ergonomically intelligent pHRI},
  author={Yazdani, Amir and Novin, Roya Sabbagh and Merryweather, Andrew and Hermans, Tucker},
  booktitle={2022 IEEE/RSJ International Conference on Intelligent Robots and Systems (IROS)},
  pages={9124--9131},
  year={2022},
  organization={IEEE}
}

@inproceedings{oleynikova2016signed,
  title={Signed distance fields: A natural representation for both mapping and planning},
  author={Oleynikova, Helen and Millane, Alexander and Taylor, Zachary and Galceran, Enric and Nieto, Juan and Siegwart, Roland},
  booktitle={RSS 2016 workshop: geometry and beyond-representations, physics, and scene understanding for robotics},
  year={2016},
  organization={University of Michigan}
}

@article{li2024configuration,
  title={Configuration space distance fields for manipulation planning},
  author={Li, Yiming and Chi, Xuemin and Razmjoo, Amirreza and Calinon, Sylvain},
  journal={arXiv preprint arXiv:2406.01137},
  year={2024}
}

@article{liu2023collision,
  title={Collision-free motion generation based on stochastic optimization and composite signed distance field networks of articulated robot},
  author={Liu, Baolin and Jiang, Gedong and Zhao, Fei and Mei, Xuesong},
  journal={IEEE Robotics and Automation Letters},
  volume={8},
  number={11},
  pages={7082--7089},
  year={2023},
  publisher={IEEE}
}

@inproceedings{marin2018optimizing,
  title={Optimizing contextual ergonomics models in human-robot interaction},
  author={Marin, Antonio Gonzales and Shourijeh, Mohammad S and Galibarov, Pavel E and Damsgaard, Michael and Fritzsch, Lars and Stulp, Freek},
  booktitle={2018 IEEE/RSJ International Conference on Intelligent Robots and Systems (IROS)},
  pages={1--9},
  year={2018},
  organization={IEEE}
}

@article{gomez2017musculoskeletal,
  title={Musculoskeletal disorders: OWAS review},
  author={G{\'o}mez-Gal{\'a}n, Marta and P{\'e}rez-Alonso, Jos{\'e} and Callej{\'o}n-Ferre, {\'A}ngel-Jes{\'u}s and L{\'o}pez-Mart{\'\i}nez, Javier},
  journal={Industrial health},
  volume={55},
  number={4},
  pages={314--337},
  year={2017},
  publisher={National Institute of Occupational Safety and Health}
}

@article{steven1995strain,
  title={The strain index: a proposed method to analyze jobs for risk of distal upper extremity disorders},
  author={Steven Moore, J and Garg, Arun},
  journal={American Industrial Hygiene Association Journal},
  volume={56},
  number={5},
  pages={443--458},
  year={1995},
  publisher={Taylor \& Francis}
}

@inproceedings{lorenzini2019new,
  title={A new overloading fatigue model for ergonomic risk assessment with application to human-robot collaboration},
  author={Lorenzini, Marta and Kim, Wansoo and De Momi, Elena and Ajoudani, Arash},
  booktitle={2019 International Conference on Robotics and Automation (ICRA)},
  pages={1962--1968},
  year={2019},
  organization={IEEE}
}

@article{peternel2019selective,
  title={A selective muscle fatigue management approach to ergonomic human-robot co-manipulation},
  author={Peternel, Luka and Fang, Cheng and Tsagarakis, Nikos and Ajoudani, Arash},
  journal={Robotics and Computer-Integrated Manufacturing},
  volume={58},
  pages={69--79},
  year={2019},
  publisher={Elsevier}
}

@inproceedings{nejadasl2022neuroergo,
  title={Neuroergo: a deep neural network method to improve postural optimization for ergonomic human-robot collaboration},
  author={Nejadasl, Atieh Merikh and Gheibi, Omid and Van de Perre, Greet and Vanderborght, Bram},
  booktitle={2022 International Conference on Robotics and Automation (ICRA)},
  pages={7372--7378},
  year={2022},
  organization={IEEE}
}

@inproceedings{li2025integrating,
  title={Integrating ergonomics and manipulability for upper limb postural
optimization in bimanual human-robot collaboration},
  author={Li, Chenzui and Chen, Yiming and Wu, Xi and Barresi, Giacinto and Chen, Fei},
  booktitle={2025 IEEE/RSJ International Conference on Intelligent Robots and Systems (IROS)},
  pages={},
  year={2025},
  organization={IEEE}
}

@article{vatsal2021biomechanical,
  title={Biomechanical motion planning for a wearable robotic forearm},
  author={Vatsal, Vighnesh and Hoffman, Guy},
  journal={IEEE Robotics and Automation Letters},
  volume={6},
  number={3},
  pages={5024--5031},
  year={2021},
  publisher={IEEE}
}

@article{kim2021human,
  title={A human-robot collaboration framework for improving ergonomics during dexterous operation of power tools},
  author={Kim, Wansoo and Peternel, Luka and Lorenzini, Marta and Babi{\v{c}}, Jan and Ajoudani, Arash},
  journal={Robotics and Computer-Integrated Manufacturing},
  volume={68},
  pages={102084},
  year={2021},
  publisher={Elsevier}
}

@article{gomes2021multi,
  title={Multi-objective trajectory optimization to improve ergonomics in human motion},
  author={Gomes, Waldez and Maurice, Pauline and Dalin, Elo{\"\i}se and Mouret, Jean-Baptiste and Ivaldi, Serena},
  journal={IEEE Robotics and Automation Letters},
  volume={7},
  number={1},
  pages={342--349},
  year={2021},
  publisher={IEEE}
}

@article{figueredo2021planning,
  title={Planning to minimize the human muscular effort during forceful human-robot collaboration},
  author={Figueredo, Luis FC and Aguiar, Rafael De Castro and Chen, Lipeng and Richards, Thomas C and Chakrabarty, Samit and Dogar, Mehmet},
  journal={ACM Transactions on Human-Robot Interaction (THRI)},
  volume={11},
  number={1},
  pages={1--27},
  year={2021},
  publisher={ACM New York, NY}
}

@article{merikh2024ergonomically,
  title={Ergonomically optimized path-planning for industrial human--robot collaboration},
  author={Merikh Nejadasl, Atieh and Achaoui, Jihad and El Makrini, Ilias and Van De Perre, Greet and Verstraten, Tom and Vanderborght, Bram},
  journal={The International Journal of Robotics Research},
  volume={43},
  number={12},
  pages={1884--1897},
  year={2024},
  publisher={SAGE Publications Sage UK: London, England}
}

@article{ovur2025optimising,
  title={Optimising Ergonomics for Robot-to-Human Object Handovers},
  author={Ovur, Ertug and Quesada, Rodrigo Chacon and Demiris, Yiannis},
  journal={IEEE Transactions on Cognitive and Developmental Systems},
  year={2025},
  publisher={IEEE}
}

\vspace{3pt}

\end{document}